# Surgery Scene Restoration for Robot Assisted Minimally Invasive Surgery

*Shahnewaz Ali, Yaqub Jonmohamadi, Ross Crawford, Davide Fontanarosa, Ajay K. Pandey*

[1]Robotics and Autonomous Systems, School of Electrical Engineering and Robotics, Queensland University of Technology, Gardens Point, Brisbane, QLD 4001, AUSTRALIA.

*Abstract*—Minimally invasive surgery (MIS) offers several advantages including minimum tissue injury and blood loss, and quick recovery time, however, it imposes some limitations on surgeon's ability. Among others such as lack of tactile or haptic feedback, poor visualization of the surgical site is one of the most acknowledged factors that exhibits several surgical drawbacks including unintentional tissue damage. To the context of robot assisted surgery, lack of frame contextual details makes vision task challenging when it comes to tracking tissue and tools, segmenting scene, and estimating pose and depth. In MIS the acquired frames are compromised by different noises and get blurred caused by motions from different sources. Moreover, when underwater environment is considered for instance knee arthroscopy, mostly visible noises and blur effects are originated from the environment, poor control on illuminations and imaging conditions. Additionally, in MIS, procedure like automatic white balancing and transformation between the raw color information to its standard RGB color space are often absent due to the hardware miniaturization. There is a high demand of an online preprocessing framework that can circumvent these drawbacks. Our proposed method is able to restore a latent clean and sharp image in standard RGB color space from its noisy, blur and raw observation in a single preprocessing stage (end-end one shot manner).

## I. Introduction

**M**inimally invasive surgery (MIS) offers several benefits, namely minimum tissue injury and blood loss, and quick recovery time. In this surgical procedure a miniaturized camera, lighting source, and tools are introduced into the patient's body through small incisions. Similarly, procedure like endoscopy uses long wired miniaturized camera and light source to view and diagnose internal organs. Recently, capsule-like endoscopes referred to as wireless capsule endoscopes (WCE) are gaining clinical attention for their ability to diagnose and deliver drugs into body. WCE uses wireless communication technology to transmit the captured video frames of internal organs. These endoscopic frames are used to take clinical decisions. With the advancement in robotics, robot-assisted MIS (RMIS) is gaining traction where visual information obtained from an endoscope can be used to the enhance surgical procedure by means of tracking tissue and tool, segmenting tissue structure for context awareness, estimating camera pose, and inferring three dimensional structure of surgical site [1-3].

Despite their current benefits, in all MIS procedures the acquired video frames experience a set of image degrading factors, therefore, information becomes compromised. To the context of RMIS, these degraded frames achieved limited success on high level vision task that can even make the whole the vision-based tasks fail [4].

Frame selection can be an alternative. However, in MIS most of the frames experience at least a minimum level of noise artifacts. It is expected that degraded frames are restored in real time from its noisy observations- referred to as image restoration (IR). This article discusses IR procedure to the context of knee arthroscopy- the MIS procedure to treat knee bone-joint ailments. However, the proposed method is also applicable to other MIS.

During the knee arthroscopy, the imaging device and tool are inserted into the knee cavity. Sterile salt water or saline is used to expand knee in order to create additional space for camera and tools. During this surgical and diagnostic procedure, the imaging device capture video sequences at close proximity to tissue structure (10-millimeter distance) [5] with 30 or 70-degree field-of-view (FoV). Here, at Queensland University of Technology (QUT), the stereo arthroscope [5, 7] has been developed that supports up to110 degree FoV. Though it increases FoV, yet at this close proximity only small portions of global scene context are accessible where frame details are compromised to several attenuation factors represents in Fig.1.

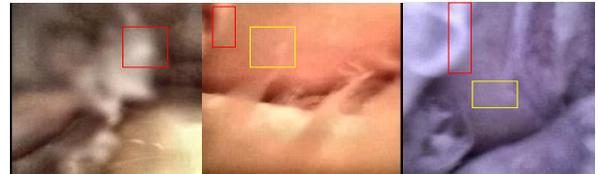

**Fig.1.** Figure represents raw arthroscopic video sequences of three different cadaver samples. Frames are obtained from muC103A camera sensor. The representative sequences are degraded by several motions blur (red rectangle) and additive noises (yellow rectangle). Due to lack of automatic white balance hardware, the acquired frames yield different color representations under halogen and white micro LED illuminants.

Like other MIS procedures, saturated pixels due to strong specular light reflections are very common to arthroscopic and endoscopic video frames [6]. Missing frame details due to saturated pixels have significant drawbacks on RMIS [7]. Other factors such as noise and blur are observed in almost all MIS frames as represented in Fig.1. In arthroscopy, substances such as body fluid and dissected particles can dissolve into water therefore, debris and hazing effects are strongly observed. The most occurring noise is gaussian noise. Other types of noise such as speckle noise, salt-pepper noise and poisson noise can be observable [8]. Impulsive noise and quantization noise can rise from hardware limitations. Moreover, debris can produce strong backscattering that can cause salt- pepper and speckle like noise. Acquired frames effectively compromised to blur by the factors like limited control on imaging devices such as exposure and aperture, unsteady hand movement (shaking), motion caused by camera steering and maneuver, and motions from floating tissue and water flow. Frame deblurring can be caused by several individual motions from different directions. It derives a hard problem to estimate accurate motion kernel and its direction adaptively which is necessary for kernel based deblurring methods [9-14]. Additionally, blur due to defocus is observed in many MIS frames which can be caused by light, specular reflection and improper focal settings. Lack of automatic white balance correction due to hardware miniaturization, MIS imaging device provides raw red, green and blue (RGB) color frames where contributions of illuminations are strongly observed which can be contaminated by different color temperature [5]. It is expected to have an automatic method that can generates white balanced standard RGB (sRGB) image which can facilitate other high-level vision tasks such as segmentation [5].

In this article the pivotal factors such as blind denoising, blind deblurring and automatic white balance are discussed, and subsequently a deep learning-based method is proposed which retrieves clear and sharp images from its blurry, noisy and raw RGB observations in real time. Due to a clean and sharp frame exhibits more image details such as features and edges, it is of a great impact on higher level vision tasks such as stereo matching [71] and segmentation [5].

## II. RELATED WORK

From the past to now, IR has been well discussed in computer vision and image processing. It is a well-defined ill-posed problem can be expressed as follows:

$$I(x_{i,j}) = G * x_{i,j} + \epsilon \quad (1)$$

Where $\epsilon$ is defined as an additive white gaussian noise (AWGN) with standard deviation σ. On the other side, $G$ is a transformation matrix that produces blur effect and I is representing resultant image where $x_{i,j}$ stands for pixel value at two-dimensional (2D) image plane. During the IR process, blur kernels are estimated, and deconvolution operation is performed over the image. Then the noise residuals are subtracted from the resultant image.

To the context of image deblurring, several methods have been proposed [9-14, 16-31]. Parameterized kernel-based methods are most widely used in the past to estimate motion blur before the learning-based method [9-14, 16-23]. Accuracy of these methods strongly depend on the definition of the motion kernels and their directions. However, the acknowledged drawbacks of kernel based methods are [32]; i.) defining an accurate kernel itself a tedious and error prone task ii.) method's accuracy are limited when noisy environment is considered iii.) often produces artifacts when kernel is not properly defined or image exhibits some sort of discontinuity. Therefore, robust, accurate, and generalization of kernel-based approach still considered as an unsolved research problem. Due to its limitations, kernel based primitive methods are becoming obsolete, and in recent years learning-based methods are gaining attentions in the computer vision community where a kernel free deblurring are achieved from unknown blur parameter such as point of spread. Method [24] used convolutional neural network (CNN) on multiscale image pyramid with their modified residual learning block named ResBlock that helps fast converging. [25] method extends the capacity of CNN to address bicubic degradation model and successively restore super resolution image from low resolution noisy input. Success also received with generative deep learning-based methods. Generative adversarial model has two networks known as generator and discriminator. Method [26] used two generative model to address maximum a posterior of deblurring method from its noisy one and retrieve blur kernel and latent image. Similarly, [27-28] methods proposed generative models with residual block that achieved state-of-art accuracy. In their implementation they used two strided convolution blocks, nine residual blocks and two transposed convolution blocks [28]. Additionally, they used $L_2$ perceptual loss and adversarial loss.

Similarly, denoising is also a long-standing IR task that has been well-studied previously. In past, denoised images are retrieved using filter-based methods. To this context, filters can be categorized by local and non-local filter [34]. Local filter uses a supporting window and statistical methods to interpolate the central pixel value. In non-local, the statistical methods are performed on several windows over the entire image for each pixel value. Gaussian [35], non-local means [36], bilateral [37], etc. are most common filters discussed in this section. These primitive filters though produce smooth images but has drawbacks of i.) vanishing weak edges and features therefore produces ii.) smooth and blurred images.

In quest of edge preserving denoiser, anisotropic [38], BM3D [39], and total variation [40] filters are proposed. Despite of their strong side of edge preserving ability, major drawbacks reported on literatures are [34]; i.) lack of textual information, and ii) staircase effect. Moreover, in some applications they failed to report satisfactory result [41]. In current research BM3D method considered as one of the state-of-the-art methods in this area.

Regardless of its ability to produce smooth images, filters are parameterized methods. Apart from blurring artifact, a major consequence of these methods is that it is not confirmed whether the filters can address different level of complex noises robustly. To this context, deep learning-based method achieved a significant research interest. In recent research, inspired by its robust perceptual and contextual accuracy among other methods, CNN is highly explored by the computer vision community. [42] method used dilated convolution with batch normalization and used ReLu activation function to extract residual noise from the noisy observations. [43] method, apart from residual learning, global residual learning strategy has been followed and they named it residual dense block. Recently autoencoder and decoder-based architecture achieved a significant progress in high level vision task i.e. segmentation. It offers precise feature extraction and localization at each scale that can facilitate mapping between noisy to clean image. [44] proposed DRUNet – a modified network on top of UNet [45] to address IR problem on half quadratic spline.

Apart from IR, during the arthroscopy, the effect of illumination causes slightly different color of images where the red or blue channel becomes dominant. It can affect the accuracy of other vision tasks [15]. Generally, this process defined by [15] as follows;

$$I_{sRGB} = f_{XYZ} \rightarrow sRGB(T_{raw} \rightarrow XYZ\ WB\ I_{raw}) \quad (2)$$

Mapping between raw RGB to sRGB as a part of color consistency has been explicitly discussed in many areas where illumination estimation was the key factor. Radiometric calibration and CNN have been used to address this issue [46-47]. Recently, method [15] is gaining the attraction by the computer vision community. In their work, this mapping function is addressed by k-nearest neighbor strategy that retrieves a color through the best matching of their nonlinear mapping function. Nevertheless, they also provide a dataset that contains 65000 pairs of images for different camera white balance settings. Some of the ground truth data is generated through the use of Adobe Camera Raw feature and rendered in photoshop.

Endoscopic image restoration has not been well studied and the progress in this sector is not significant. A scarce literature currently establishes IR problem [48-56] where most of the articles address specular removal, parameterized deblur, desmoke, colorization and quality assessment. Robust denoiser and deblurring mechanism in real time still remains an unsolved problem which has a countless demand for robotic vision task such as tracking and navigating robot in RMIS environment. More specifically, in arthroscopy, IR exhibits additional complexity considering the factors such as underwater environment, lack of control on imaging devices, poor imaging condition, lens distortion, debris and hazing and complex motion which requires a more sophisticated and robust solution. To facilitate high level vision task, in this article, we proposed a single framework where raw arthroscopic images are enhanced through the color correction as well as the latent clean and sharp frames are restored after performing denoising and deblur task together which is irrespective to noise level.

## III. METHODOLOGY

Restoration white balanced latent clean and sharp image $y$ from its noisy and raw sensor observation, $x$ considered as a mapping function such that;

$$y \rightarrow f(x, \theta) \quad (3)$$

Here, $f$ is the mapping function and $\theta$ are the parameters to learn during the training. In this article, this problem considered as a regression problem which achieved end-end solution for multi-IR tasks.

### A. Model

With the success of high-level vision task by UNet, this regression problem is addressed within UNet architecture.

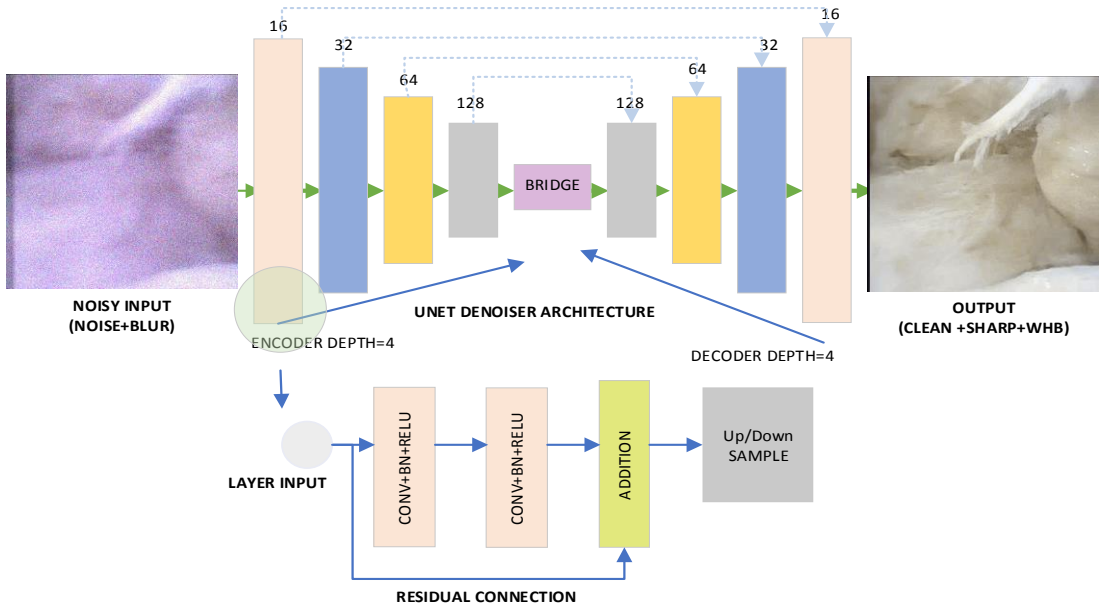

**Fig.2.** Architecture for endoscopic image restoration framework. A clean, sharp and white balanced (WHB) video frame is retrieved from its raw, noisy and blurred observation. The network depth for encoder and decoder is **4**. Network use residual connection as it is shown in bottom image. Accumulated loss function calculated from PSNR, SSIM, perception loss and reduced mean of edge loss between noisy and clean observation.

UNet is a well-known network architecture widely used for segmenting medical data as an end-to-end solution. In recent work, this architecture is revisited to address IR task, for instance DRUNet [44]. In this work, UNet architecture is followed to address color mapping and two mainstream IR tasks namely denoising and deblurring in endoscopic standpoint. Instead of classic UNet, this implementation uses of residual blocks into UNet architecture. Following the encoder-decoder manner, the contraction path of UNet precisely extract features at different scale down-sampled at each step. On the other side, the expansion path learns to localize each feature at different level up sampled at each step.

Residual learning [57] strategy, has several benefits across the network, it increases training and prediction accuracy even with small network depth. In UNet architecture spatial information loss is caused by down sampling in contraction path, and it has been confirmed that, after its name ResUnet [58,59], a residual learning strategy performs better over classic UNet. In recent works on IR [60] network such as DnCNN gets benefit from residual learning strategy. In this work, we followed the strategy of ResUnet.

Unet architecture consist of three basic building blocks, namely; encoder, decoder and connecting block. Encoder block learns high level features to its complex low-level feature representation. In this way UNet encoder learns coarse pixel-wise feature representation of raw images. When residual blocks are implemented on UNet encoder, it provides more spatial information that means more noisy spatial representations are obtained. Similarly, on the decoder part, UNet learns pixel-wise fine feature from its coarse complex representation, for instance blur weak edges to its sharp representation. It subsequently preserves contextual information thus produces clean and sharp image in end-to-end fashion. Batch normalization is widely recognized for faster training when input distributions are different known as an internal covariate shift. [42,60] methods received benefits with the use of batch normalization to learn noisy residual image. Noise such as gaussian and others can have different statistical distributions around the arthroscopic sequences. Moreover, blur can occur at different level from multiple motions a well observed occurrence in underwater MIS. It is understood that, these motions can exhibit blur effect at different directions in images, even a single pixel can experience several motions having different motion directions. To incorporate level independent noises including blur effect which means diverse input distributions, batch normalization strategy has been followed.

*B. Dataset and Training*

Arthroscopic video sequences are recorded during the knee arthroscopy of cadaver samples. Total five knee arthroscopy has been conducted with five cadaver knee samples. During these experiments some frames are captured at steady camera positions. Lighting conditions are maintained by the use of manually adjusted illumination controller. However, a little contribution of motion blurs and defocusing are observed at some distant part which are not considered and when possible are corrected using the methods [27, 30, 61]. White balanced are obtained from the [15] method and corrected through the raw image editing tools when it was required. Corrected color values are validated by reconstructing its reflectance and compared with spectrometer data as mentioned by the [62] method. Clean images are then degraded by adding multi-level of Gaussian, Speckle, Salt and Pepper, and Poisson noise. Blur images are generated through the use of motion blur kernel. Along with these we used 400 image samples from [33] and the whole dataset is splitted into three categories, i.) clean image ii.) blur image iii.) noisy and blur image.

Unet architecture also learns to localize its feature from local scope to global. To facilitates this to the context of arthroscopy,

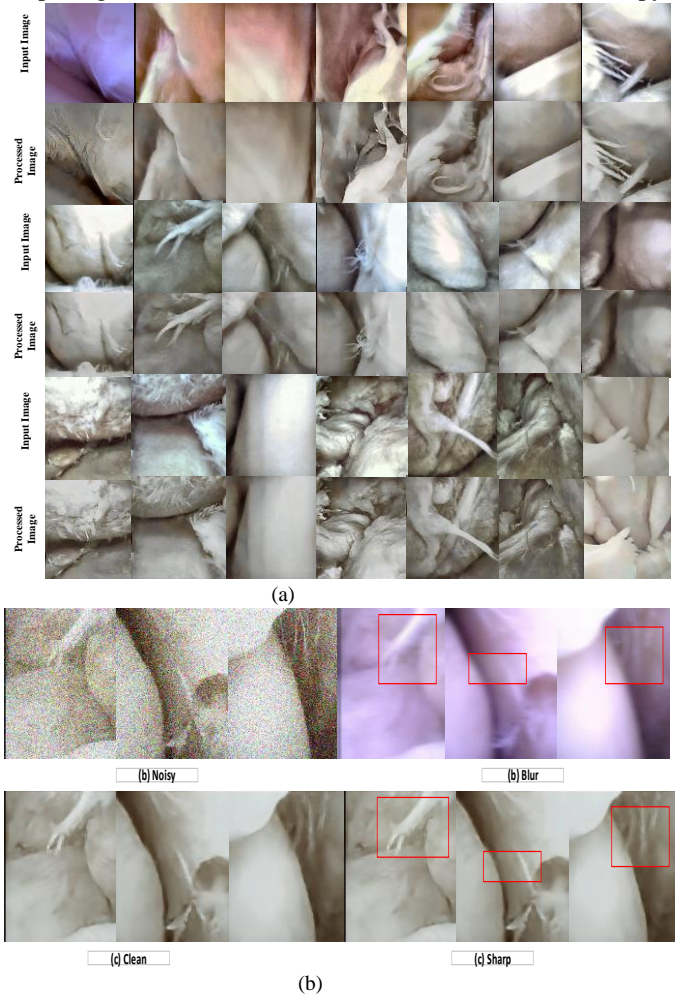

(a)

(b) Noisy     (b) Blur

(c) Clean     (c) Sharp

(b)

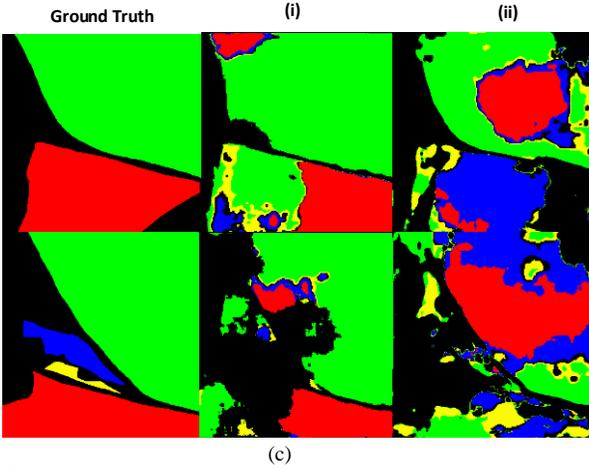

**Fig.3.** Figure-(a) represents the visual representation of real scene and the result obtained from our method. First row represents real arthroscopic scene and the second row represents the results and so on. Figure-(b) represents the outcome of IR tasks considering high level noisy and blur data. (c) represents arthroscopic scene segmentation results. Left column represent ground-truth label, column (i) represents segmentation results obtained from preprocessed dataset using our method and column (ii) represents result obtained from same dataset without preprocessing. It is clearly showing that this framework increases the accuracy of segmentation task.

we also use synthetic rendered arthroscopic video sequences using 3D graphics software-blender. The training has been performed on 4500 images and validated against 1500 images. During the test we use total 6,803 arthroscopic video frames from all five cadaver samples.

During the training structural similarity index (SSIM), peak signal to noise ratio (PSNR), perception loss- $L_2$ norm, and loss of edges between noisy and clean observations are evaluated individually. Rather than individual loss function, it has been found that with the use of accumulated loss function of SSIM, PSNR and $L_2$ norm the network converged smoothly and obtained better validation and test accuracy. Hence, the total loss defined as,

$$Loss_{total} = \sum(L_{SSIM} + L_{PSNR} + L_2) \quad (4)$$

Loss $L_{PSNR}$ and $L_2$ norm are used to define the network learning strategy to reconstruct clean image from its noisy observation as well as the color mapping function. PSNR is defined as follows [63];

$$PSNR = 10 * \log 10 \left(\frac{max^2}{MSE}\right) \quad (5)$$

Where,

$$MSE = \frac{1}{M*N*O} \sum_{x=1}^{M} \sum_{y=1}^{N} \sum_{z=1}^{O} \left[(I_{(x,y,z)} - I'_{(x,y,z)})^2\right] \quad (6)$$

$max$ is the highest value of gray scale image. Similarly, SSIM and difference of edges between a noisy and blur image are used for deblurring during the fine-tuned training stage. In this following strategy, network learns sharp edges and features from its coarse blur representations.

Learning noises comparatively simpler task than deblurring. Moreover, deblurring under noisy condition is relative complex task than the straightforward deblurring. We followed two stage training procedure, i.) coarse training, and ii.) fine-tuned training. The network outperforms, when it is trained with all noisy and blur observations for coarse training and then fine-tuned training is done over blur dataset.

## IV. RESULT

Arthroscopic video sequences are evaluated against several state-of-the-art methods. Dataset contains noisy observations, named as real and its corresponding clean images act as ground truth data. Real data set contains raw observation of arthroscopic scenes which are compromised to noise and blur. Moreover, these frames are not white balanced, provides raw RGB color. It is worthwhile to mentioned that, many frames perceptually exhibit small contextual information due to lighting condition are not uniform inside the knee cavity, therefore, frame contains both saturation and underexposed image parts. Additive white gaussian noise (AWGN) are added to the raw input frames with standard deviation 25. To simulate debris, haze and random back scattering noise like speckle, salt-pepper and poisson are added to the real video frames. Additionally, to achieve several level of blurring effects, both real and raw image are convolved with blur kernels. To compare denoising result, the state-of-the-art algorithms, including, Gaussian [35], non-local mean filter [36], Bilateral filter [37], Anisotropic filter [38], BM3D [39], For deblurring BM3D-deblur [39], Bayesian-based iterative [64], unsupervised wiener [65], l0 gradient prior [66], Total variation deconvolution [67], natural image statistics [68], deblurring under high noise levels [69] and deep learning based method, deep CNN denoiser prior [42], Deblurgan [28], Scale-recurrent network [30] are evaluated.

TABLE I
COMPARISON OF IR TASK ON ARTHROSCOPIC FRAMES

| ID | Gaussian Noise | | | | | | | | | | | | Deblur | |
| --- | --- | --- | --- | --- | --- | --- | --- | --- | --- | --- | --- | --- | --- | --- |
| | $\sigma = 10$ | | $\sigma = 20$ | | $\sigma = 30$ | | $\sigma = 40$ | | $\sigma = 50$ | | $\sigma = 60$ | | SSIM | PSNR |
| | SSIM | PSNR | SSIM | PSNR | SSIM | PSNR | SSIM | PSNR | SSIM | PSNR | SSIM | PSNR | | |
| GAS[35] | 0.936 | 31.16 | 0.84 | 30.26 | 0.726 | 29.04 | 0.626 | 27.7 | 0.536 | 26.36 | 0.465 | 25.09 | N/A | |
| NLM[36] | 0.942 | 38.7 | 0.908 | 35.7 | 0.771 | 29.80 | 0.512 | 24.03 | 0.310 | 20.13 | 0.19 | 17.0 | N/A | |
| BILT[37] | **0.960** | **41.36** | 0.89 | 37.12 | 0.670 | 31.48 | 0.410 | 26.7 | 0.245 | 23.19 | 0.157 | 20.6 | N/A | |
| AISO[38] | 0.782 | 27.5 | 0.917 | 32.8 | 0.917 | 28.7 | 0.32 | 24.0 | 0.16 | 24 | 0.106 | 18.2 | N/A | |
| BMD[39] | 0.936 | 30.8 | 0.932 | 30.61 | **0.925** | **30.13** | **0.914** | **29.3** | **0.899** | **28.29** | **0.899** | **29.5** | N/A | |
| BMD[39] deblur | 0.898 | 22.59 | 0.896 | 22.93 | 0.893 | 23.5 | 0.885 | 23.8 | 0.870 | 24.03 | 0.844 | 24.1 | 0.85 | 21.2 |
| IRCN[42] | 0.956 | 40.0 | **0.956** | **39.0** | 0.819 | 33.29 | 0.428 | 25.92 | 0.231 | 22.14 | 0.141 | 19.6 | 0.89 | 19.8 |
| GAN[28] | 0.65 | 27.5 | 0.45 | 25.2 | 0.36 | 23.9 | 0.30 | 22.9 | 0.269 | 22.09 | 0.24 | 21.28 | 0.85 | 25.2 |
| SRN[30] | 0.62 | 30.26 | 0.37 | 26.3 | 0.23 | 23.13 | 0.16 | 20.8 | 0.12 | 19.17 | 0.1 | 17.8 | **0.92** | **27.5** |
| LCY.[64] | N/A | | | | | | | | | | | | 0.90 | 27.5 |
| WIN.[65] | N/A | | | | | | | | | | | | 0.80 | 32.2 |
| 10GR[66] | N/A | | | | | | | | | | | | 0.82 | 29.0 |
| TV[67] | N/A | | | | | | | | | | | | 0.83 | 24.5 |
| NI[68] | N/A | | | | | | | | | | | | 0.87 | 21.6 |
| HN[69] | N/A | | | | | | | | | | | | 0.76 | 21.1 |
| OUR | **0.93** | **34.1** | **0.94** | **35.0** | **0.93** | **33.9** | **0.92** | **31.9** | **0.82** | **29.2** | **0.61** | **25.82** | **0.94** | **37** |

| | Speckle, Salt Pepper, Poisson Noises + Blur | |
| --- | --- | --- |
| | SSIM | PSNR |
| GAS[35] | 0.825073 | 26.783526 |
| NLM[36] | 0.803788 | 30.834656 |
| BILT[37] | 0.792327 | 31.219724 |
| AISO[38] | 0.730226 | 25.649356 |
| BMD[39] | **0.865542** | **26.563407** |
| BMD[39] deblur | 0.825373 | 20.884970 |
| IRCN[42] | 0.832018 | 31.624574 |
| GAN[28] | 0.736981 | 24.844043 |
| SRN[30] | 0.756599 | 30.071061 |
| LCY.[64] | 0.765477 | 32.147631 |
| WIN.[65] | N/A | |
| 10GR[66] | N/A | |
| TV[67] | N/A | |
| NI[68] | N/A | |
| HN[69] | N/A | |
| OUR | **0.84** | **30.63** |

The quantitative representations are expressed using metrics average PSNR and SSIM in Table I. Perceptual representations are presented in Fig.3. To demonstrate the impact of our method on high level vision task, arthroscopic scene segmentation is performed. The same neural network used by method [5] is trained for this task using both raw and preprocessed data using this method. On the same test set, the accuracy improvement for Femur, Anterior Cruciate Ligament (ACL), Tibia, Meniscus are 2.6%, 2% 6.3% and 7% (Fig.3).

## V. CONCLUSION

In this work, image restoration framework for knee arthroscopy is presented. Visual limitations exhibit great challenges for high level vison task such as stereo matching [71]. Clean and sharp video frames are more informative. It is confirmed from the obtained result that, this framework restored clean and enhanced frames consist of more textual information.

The framework is established following the encoder-decoder
6

like convolutional neural network architecture -Unet. Strategy like Residual learning and batch normalization has been followed that speed up training phase. The resultant network obtained highest accuracy when perceptual loss, PSNR, SSIM and edge difference loss are summed up referred as total loss in two stage training. Training phases used Adam optimizer with learning rate 1e-4. It takes 0.024 seconds to process each frame with the use of Nvdia telsa-P100 GPU.

ACKNOWLEDGMENT

This work is supported by Australian Indian Strategic Research Fund Project AISRF53820 and in part by Australian Centre for Robotic Vision. This work used AISRF53820 knee arthroscopy dataset of cadaver samples.